\documentclass[preprint,12pt]{article}
\usepackage{amssymb}
\usepackage{comment}
\usepackage{wrapfig} 
\usepackage{graphicx,amssymb}
\usepackage{color}
\usepackage[numbers]{natbib}
\usepackage{multirow}
\usepackage{subfig}
\usepackage{colortbl}
\usepackage[binary-units=true]{siunitx}
\usepackage{breqn}
\usepackage{authblk}
\usepackage{array}
\begin{document}

\title{Within-Brain Classification for Brain Tumor Segmentation}




\date{\vspace{-5ex}}
\author[1]{Mohammad Havaei \thanks{mohammad.havaei@gmail.com}}

\author[1,2]{Hugo~Larochelle\thanks{hugo.larochelle@usherbrooke.ca}}
\author[1]{Philippe Poulin\thanks{philippe.Poulin2@usherbrooke.ca}}

\author[1,3]{Pierre-Marc Jodoin\thanks{pierre-marc.jodoin@usherbrooke.ca}}

\affil[1]{Universit\'e de Sherbrooke, Sherbrooke, Canada}
\affil[2]{Twitter, USA}
\affil[3]{Imeka, Canada}




\maketitle

\begin{abstract}
Purpose:
In this paper, we investigate a framework for interactive brain tumor segmentation which, at its core, treats the problem of interactive brain tumor segmentation as a machine learning problem. 

Methods:
This method has an advantage over typical machine learning methods for this task where generalization is made across brains. The problem with these methods is that they need to deal with intensity bias correction and other MRI-specific noise. In this paper, we avoid these issues by approaching the problem as one of {\it within brain generalization}. Specifically, we propose a semi-automatic method that segments a  brain tumor by training and generalizing within that brain only, based on some minimum user interaction. 

Conclusion:
We investigate how adding spatial feature coordinates (i.e. $i$, $j$, $k$) to the intensity features can significantly improve the performance of different classification methods such as SVM, kNN and random forests. This would only be possible within an interactive framework.  We also investigate the use of a more appropriate kernel and the adaptation of hyper-parameters specifically for each brain. 

Results:
As a result of these experiments, we obtain an interactive method whose results reported on the MICCAI-BRATS 2013 dataset are the second most accurate compared to published methods, while using significantly less memory and processing power than most state-of-the-art methods.
\end{abstract}

\section{Introduction}
Brain tumor segmentation is primarily used for diagnosis, patient monitoring, treatment planning, neurosurgery planning and radiotherapy planning. The task of brain tumor segmentation is to locate the tumor and delineate different sub-regions of the tumor, namely {\em edema, non-enhanced}, and {\em enhanced} regions (see Fig. 1).  A standard way to diagnose a brain tumor is by using magnetic resonance imaging (MRI), for which many different modalities can be used. The most frequent MRI modalities used for brain tumor segmentation are Flair, T1-weighted (also referred to as T1), T2- weighted (also referred to as T2) and T1-weighted contrast-enhanced (gadolinium-DTPA) which we refer to as T1C. These different modalities are often used jointly as they provide complementary information for locating tumors.

Unfortunately, tumors (especially glioblastomas and metastases) can appear almost anywhere in the brain. They have no prior shape, and often have poorly defined edges.  Also, they visually present themselves in grayscales that are present in healthy tissues as well.  As a consequence,  brain tumor segmentation in practice is still done manually.  Manual segmentation is not only time consuming and tedious, it is also subject to variations between observers and also within the same observer \cite{schmidt2005}.

\begin{wrapfigure}{r}{0.4\textwidth}

	\centering \includegraphics[width=0.9\linewidth]{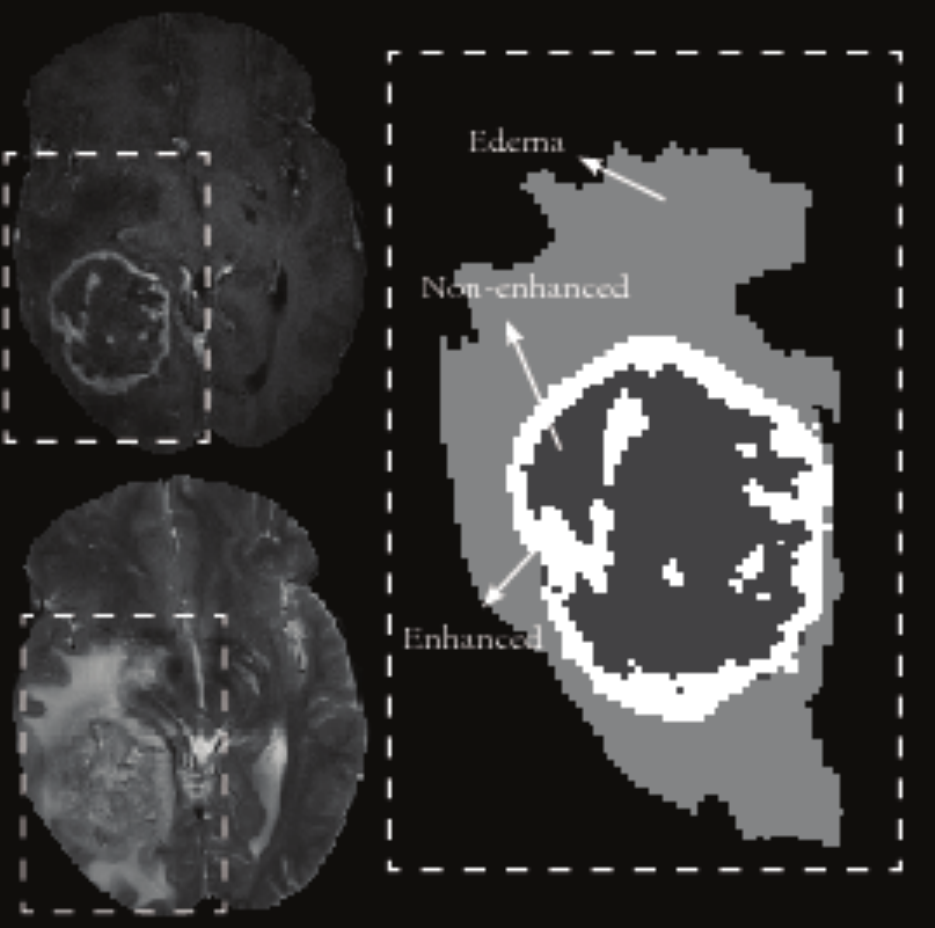}
	\caption{\footnotesize {\bf Left:} T1C and T2 modality. {\bf Right:} groundtruth tumor segmentation.}
	
	\label{fig:brain_seg} 
 
\end{wrapfigure}

Many methods have been proposed to facilitate the tumor segmentation process. Among them, {\it automatic} methods, which rely on machine learning, are very popular and in some cases very efficient \cite{bauer2013}. These methods are trained on a number of subjects and generalize on data which might be gathered from different MRI scanners. Because there is no intensity standardization among MRI scanners, this makes generalization difficult for automatic methods. In an attempt to overcome these difficulties,  a lot of prepossessing steps are made which can be time consuming. Also, to improve generalization, these methods often compute high dimensional feature vectors~\cite{schmidt2005} which add to the processing time and take up a lot of memory.

In this paper, we consider the specific problem of segmenting an imaged brain into 4 classes: edema, non-enhancing tumor, enhancing tumor and healthy tissue (see Fig.~\ref{fig:brain_seg}).  Note that the non-enhancing tumor sometimes includes necrotic tissue. Our approach is halfway between automatic and semi-automatic methods. While machine learning methods train on a pre-selected set of brains and then generalize to testing brains, our method implements a ``single brain" supervised learning method.  The user roughly selects brain voxels associated to each class and then these voxels are used as training data.  The method then generalizes by labeling non-selected voxels.  

The main characteristics of our method are as follows:
\begin{itemize}
  \item Since it treats each brain as a separate dataset, it is immune to the multi-MRI disadvantages mentioned above.
  \item Although it uses only 6 simple features, it produces highly accurate results.
  \item The segmentation process for a $ 240\times 240 \times 168$ brain takes approximately 10 seconds for our fastest method which is much faster than most state-of-the-art methods which can take up to 100 minutes. 
  \item The method is extremely memory efficient (50 Mb vs. \ $>$2 Gb for other methods)
\end{itemize}

In this paper, we first evaluate the performance of a $k$ nearest neighbor classifier (kNN) within this framework. Then, we extend this framework and thoroughly evaluate its potential through comparing the use of several classifiers, including support vector machines (SVM), random forests and boosted decision trees. Second, we propose better distance metrics to be used by SVM classifier in the context of this approach. We also investigate the importance of performing hyper-parameter selection individually for each brain, as opposed to using generic hyper-parameters for every brain. Thanks to this investigation, we were able to significantly improve the resulting brain segmentation system and achieve a competitive performance compared to the methods submitted to the brain tumor segmentation challenge (BRATSURL~\cite{BRATSURL}) online evaluation benchmark.

\section{Related Work}

Brain tumor segmentation methods can be divided into {\it automatic} methods and {\it semi-automatic} (interactive) methods.
Semi-automatic methods are those relying on user interaction. Most of these methods use either deformable models or classification methods to perform segmentation (see \citet{bauer2013} for a survey).

For automatic methods, machine learning classification techniques are a tool of choice for designing such systems, as they can easily integrate different MRI modalities as well as other features.  After integrating different intensity and texture features, these methods decide to which class each voxel belongs to.

For instance, Festa et al.~\citep{Menze2014} used a series of intensity and texture based features to make a feature space of over 300 dimensions, on which a random forest classifier was trained. Tustison et al.\ and Reza et al.\ also used random forests~\citep{Menze2014}. Tustison et al.\ constructed a multi-dimensional feature space by incorporating first order neighborhood statistical images, GMM and Markov Random Field (MRF) posteriors, and template differences.
\cite{lee2008} performed binary segmentation (tumor vs.\ non-tumor) using T1, T2, T1C in an SVM framework followed by a variation of conditional random fields to account for neighborhood relationships.
\cite{bauer2011} used a kernel SVM for multiclass segmentation of brain tumors, where a CRF is used to regularize the results.

\citet{schmidt2005} compared the combination of many different feature sets, such as binary mask, average intensity, left to right symmetry.
\citet{luts2007} also compared different feature selection methods such as Fisher discriminant analysis, Kruskal wallis, relief-f and ARD for LS-SVM. 

Because automatic methods train on multiple brains, these methods are vulnerable to the variations in the MRI data. These variations come from the fact that MR images are generated by different machines and each have their own unique noise and intensity level. To overcome this difficulty, most of these methods rely on a large number of features, which requires a lot of memory and computation time. 

As for semi-automatic methods, deformable models are often employed. These algorithms are usually initialized by a user drawing a contour around the tumor. Following an energy minimization criterion, the contour shrinks down towards the borders of the tumor~\cite{Jiang2004,Wang2009}. \citet{hamamci2012a} used a so-called CA-based method on T1 weighted images to produce a probability map for the tumor, based on seeds provided by the user. This probability map is later used in a level set framework. Later, they extend their method to accept multi-modal MRI inputs namely T1C and Flair. For a two class segmentation (tumor, edema) this method takes 1 minute for user interaction and 10-20 minutes for segmentation depending on the size of the tumor~\cite{hamamci2012b}.
There exists a line of research focusing on how to efficiently initialize the active contour and thus remove user interaction. In this context, the location of the tumor is roughly determined by some other method and deformable models are used as post-processing for refinement.  
\citet{ho2002} use the difference between T1 and T1C together with a Gaussian mixture model (GMM) to get a probability map of the tumor, which is used in a level-set model to initialize the contour. \citet{prastawa2003b} used voxel registration with an atlas as a way to get a probability map for abnormalities. An active contour is then initialized using this probability map and iterates until the change in posterior probability is below a certain threshold. 

Although deformable models have been popular in medical image analysis, they have some significant disadvantages. Because these methods rely on image gradients, they are likely to fail when the object of interest does not have well defined borders. The contour may get attracted by strong gradients from  surrounding objects. Incorporating different features into the model is also non-trivial. Finally, without a GPU implementation, these methods can be extremely slow.

There has been research on ensembling results from multiple methods applied to brain tumor segmentation. \citet{huo2013} used three segmentation methods: fuzzy connectedness, GrowCut and voxel classification using SVM to generate candidate segmentations for each voxel. Confidance-based averaging (CMA) was used to make the ensemble. 

Although our method is a semi-automatic method, it shares with automatic methods the use of a machine learning classification algorithm, ran on a feature representation of voxels and improved by a spatial dependency model. The main difference is that generalization is performed {\it within} each brain, based on the training data  provided by the user's interaction. This simplified generalization problem allows us to use a very simple feature space, yielding an interactive segmentation method that is fast and effective. ~\cite{vaidyanathan1995} used a similar, semi-automatic, kNN classification method, applied to proton density, T1 and T2 modalities.
~\cite{cai2007} also proposed a semi-automatic segmentation method that uses instead  Quadratic Discriminative Aanalysis to perform multi-class segmentation. However, they did not use the $\langle i,j,k\rangle$ voxel positions as features (see Section~\ref{seq:KNN}) nor did they deal with label spatial dependency modeling (see Section~\ref{sec::CRF}), which we found to play a crucial role in obtaining competitive performances.

\section{Investigating Within-Brain Generalization} 
\label{sec:framework}

Within-brain generalization treats the segmentation of each brain as its own machine learning experiment, in which a classifier is trained (on user-labeled voxels) and used to generalize to new observations (voxels not labeled by the user). 

This approach is motivated by the observation that, with current computers and for relatively small data sets with small feature spaces, a machine learning experiment (including hyper-parameter selection) can actually be performed within a very short delay, even for more sophisticated algorithms that require more than simply storing the data (as in kNN). Moreover, segmenting only within a given brain removes the challenging problem of generalizing across brain imaging acquisition conditions. 

In what follows, we describe the details of our approach and enumerate the different variations we explored in this direction.

\begin{figure*}[h]
    \begin{center}
        \includegraphics[width=0.9\linewidth]{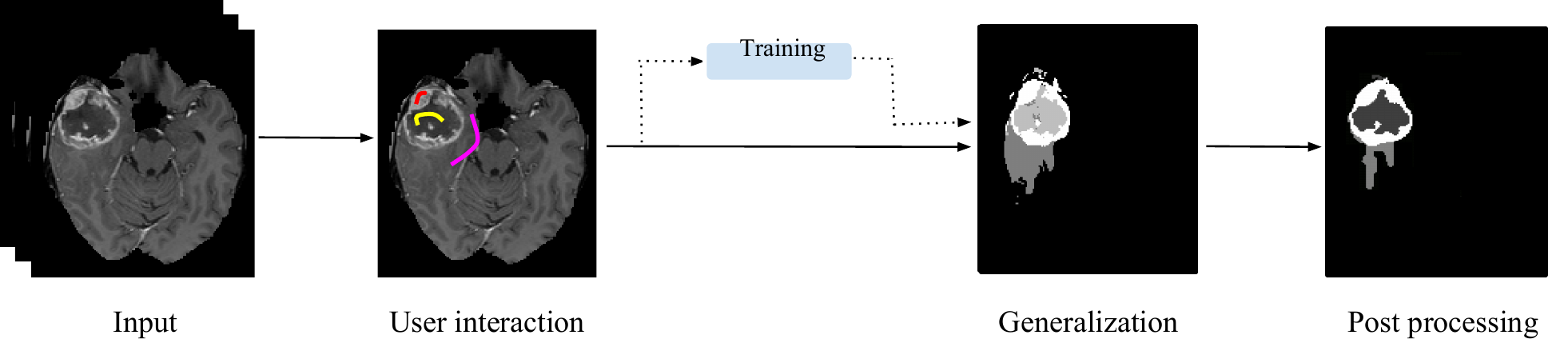}
        \vspace{-0.1cm}
	    \caption{Our method in a nutshell. The segmentation is performed on the entire brain based on data provided by user interaction.
	        }
	    \label{fig:keyidea} 
	\end{center}
\end{figure*}
Figure \ref{fig:keyidea} shows our method in a nutshell. We explain these steps in Section~\ref{sec:framework}.

\subsection{Feature representation and manual selection}

The first step of our method is to collect voxel label data for a given brain image to segment.  This is done by the user who roughly selects a subset of voxels associated with each class, through a graphical interface. The number of strokes required for obtaining the training data depends on the number of tumors in a given brain. However, usually one or two strokes per-class is enough. We will note as $B$ a binary mask such that $B_v\in \{0,1\}$ indicates whether a voxel $v$ has been manually selected ({\it i.e.} labeled) or not. $T$ will then be the class-selection mask where $T_v\in \{$edema, non-enhancing tumor, enhancing tumor, healthy$\}$ is the class label associated with the voxel $v$ by the user. 

We must also decide on a feature representation for the different voxels. Each brain image $I$ is assumed to come with 3 MRI modalities (T1C, T2, Flair), such that $I$ is a tensor where each voxel $v$ in $I$ is a 3D vector  containing the grayscale values of the modalities. This is represented by $I^1_{v},I^2_{v},I^3_{v}$. By converting each voxel $v$ to an N-dimensional feature representation $F_v$, it will be possible to train a classifier to predict the voxel label $T_v$, for every voxel, from its feature representation. We propose a simple 6 dimensional feature represeentation, which consists of the MRI modality gray scales and the 3d position of voxel $v$: $F_v = (I^1_{v},I^2_{v},I^3_{v},i,j,k)$. These features are normalized between zero and one. 




At this point, from each labeled voxel, we can thus generate a training pair $(F_v,T_v)$ and construct a training set ${\cal D}$ that we shall use to classify the non-selected voxels using a classifier.

\subsection{Voxel classifiers}
\label{sec:classifiers}

Having built the training set through manual interaction, the next step is to train a classifier and generalize the segmentation to non-selected voxels. We investigate the use of different machine learning algorithms to produce a classifier. While we could, theoretically, consider any existing algorithm, it is natural to prefer algorithms that are known to be robust and fairly "black box" in their use. For instance, we do not want the user (typically a doctor or a neuro-scientist) to have to manually tune hyper-parameters for each brain, with trial and error.  So we chose algorithms that are known to be easily tuned or for which default values of their hyper-parameters tend to work well.  These algorithms have also shown to be successful for automatic brain tumor segmentation~\cite{schmidt2005,Menze2014}.

\subsubsection{K-Nearest Neighbors (kNN)}
\label{seq:KNN}

To start, $k$ nearest neighbor (kNN), one of the simplest classifiers, is considered. For every voxel $v$, kNN finds among the training data ${\cal D}$, the set of $k$ nearest neighbors (${\cal N}_v$)  based on $F_v$. Let ${\cal N}_v=((F_{v_1},T_{v_1}),(F_{v_2},T_{v_2}),...,(F_{v_k},T_{v_k}))$ where $ F_{v_i}$ is the $i^{\rm th}$ closest training point of $F_v$.  The kNN classification rule assigns a class label to some voxel $v$ following this equation
\begin{equation}
    T_v = \arg\max_c \frac{1}{k}\sum_{(F_{v_i},T_{v_i})\in {\cal N}_v} \delta(T_{v_i},c)
\end{equation}
where $c$ is a class label and $\delta(a,b)$ returns 1 when $a=b$ and 0 otherwise. Note that this formulation can be seen as using a posterior class probability:
\begin{equation}
    p(T_v=c | F_v) = \frac{1}{k}\sum_{(F_{v_i},T_{v_i})\in {\cal N}_v} \delta(T_{v_i},c)
    \label{eq:postKNN}
\end{equation}
which states that the probability of an observation $F_v$ of being in class $c$ is given by the proportion of nearest neighbors assigned to that class.  This probabilistic formulation of the classifier will be reused for the  unary terms of a CRF, described in Section~\ref{sec::CRF}.


\subsubsection{Support Vector Machine}

The support vector machine (SVM)~\cite{cortes1995} is probably the most frequently used classifier. This is in part due to the existence of many freely available, mature and easy-to-use implementations. In its parametric form, it is a linear classifier that attempts to classify data points by maximizing the margin between the decision boundaries of the different classes and their closest points. 

Of higher interest in our setting is the kernelized version of SVM~\cite{Lampert09}. A choice for the kernel that often proves successful is the radial basis function (RBF) kernel:
\begin{equation}
\mathcal{K}( F_j , F_v ) = \exp(\mbox{-}\gamma \parallel  F_j - F_v \parallel^2_2 ).
\end{equation}
where $\gamma$ is a hyper-parameter. Also, a slack variable $C$ is used to relax the constraints in the SVM optimization problem~\cite{Lampert09}. The resulting classifier effectively takes the form of a template matcher, that compares a given input with all training examples, each voting for their class with a weight related to their similarity with the input (as modeled by the kernel). In this sense, it is similar to the kNN classifier, though the former often outperforms the later in practice. 

It is also possible to obtain a posterior class probability $p(T_v=c | F_v)$ from the SVM. This is done by training the parameters of an additional sigmoid function of the form
\begin{equation}
  P(T_v=c|F_v) = \frac{1}{1+\exp \left( Af(F_v,c)+B\right)}
  \label{eq:postSVM}
\end{equation}
where $f(F_v,c)$ is the unthresholded output of the SVM and $A,B$ are the parameters to be estimated \cite{platt1999}.  Here again, the posterior probability function will be used later on, for the CRF unary term.


%
\subsubsection{Ensemble of Decision Trees}

Another popular approach to classification are ensembles of decision trees.  Each decision tree is trained by recursively partitioning the feature space, according to some heuristic that favors a good separation of classes. Once a criterion for stopping the tree growth is reached, a conditional class distribution is then computed at each leaf, based on the training data falling into the corresponding partition. Specifically, the class distribution $p(T_v=c|F_v)$ is set as 
\begin{equation}
    P(T_v=c|F_v) = \frac{N_c}{N}
    \label{eq:postDT}
\end{equation}
where $N_c$ is the relative frequency of examples belonging to class $c$ of the partition in which $F_v$ falls and $N$ is the total number of examples.

The performance of a single decision tree is often disappointing. However, by constructing an ensemble of such trees, a competitive classification performance is achievable. There are different approaches to combining decision trees into an ensemble. The two most popular algorithms for ensembles of decision trees are random forests and Adaboost~\cite{murphy2012}.  We considered these two algorithms for our experiments.


\subsection{Distance Metric/Kernel}
\label{sec:distance}

The performances of the SVM classifier often depends on the choice of metric or kernel used to compare data points. Thus, it is generally beneficial to adapt this choice to each individual problem.  For example, the conventional RBF kernel puts equal weight to each dimension of the feature space. However, in our within-brain framework, the spatial  coordinate features $\langle i,j,k \rangle$ and the modality features actually play different roles. Intuitively, one role of the spatial coordinates is to avoid that a user-labeled voxel starts influencing the prediction made at a voxel far away from it, e.g.\ to avoid false positives in faraway regions. The modality features, are thus mostly informative within the vicinity of a user-labeled voxel.


Therefore, we might want to weight the modality and spatial features differently, within the RBF kernel of the SVM. To maintain positive-semidefiniteness of the kernel, we simply opt for using two different values of $\gamma$ for MRI modality intensities and the spatial features:
%
%
\begin{eqnarray}
    \mathcal{K}( F_j , F_v ) = \mbox{exp}(&-\gamma_1& \parallel  F_{j,{\{1:N\}}} - F_{v,{\{1:N\}}} \parallel^2_2 \\ \nonumber
        &-\gamma_2& \parallel  F_{j,{\{N+1:N+3\}}} - F_{v,{\{N+1:N+3\}}} \parallel^2_2). \label{eq:kernel-non-lin}
\end{eqnarray}
This kernel is also equivalent to the product of two RBF kernels, each defined on the subspace of modalities and of spatial coordinates, and each having their own hyper-parameters. The  hyper-parameters required by this approach are $\gamma_1$ and $\gamma_2$.

\subsection{Importance of Within-Brain Hyper-Parameter Selection}
 
When training a classifier, hyper-parameter values must be specified. One approach which is commonly implemented~\cite{Menze2014} is to choose hyper-parameters by cross-validation in a grid search approach on a subset of brains and fix the selected set of hyper-parameters for the rest of the brains. We hypothesize given the variations in MRI data, using a fixed set of hyper-parameters for generalization is not optimal. An alternative way is to  perform hyper-parameter selection individually for each brain, in order to adapt to the specificity of each case. We measure the potential gains of this approach in our experiments when selecting the hyper-parameters for the SVM, namely the slack variable $C$ and the coefficient $\gamma$.  A detailed discussion of this experiment is presented in section \ref{sec:hyper-parameter_selection}.

\subsubsection{Conditional Random Fields (CRF)}
\label{sec::CRF}

As mentioned earlier, segmentation accuracy can easily be improved by leveraging a model of the 3D spatial regularity of labels. One way of enforcing spacial regularity is to define a joint (conditional) distribution over the labels of
all voxels in the brain that expresses the expected dependencies between neighboring voxels.
Conditional Random Fields (CRF) provide a convenient formalism for that. CRFs model directly the posterior probabilities of the labels given the features $P(T|F)$ directly, alleviating the need to model the distribution over the feature vectors $F$ and allowing us to construct rich conditionals $P(T|F)$.  

Formally speaking, we use the following form for  $P(T|F)$:
\begin{equation}
    P(T|F)=\frac{1}{Z}\prod_v \phi(F_v,T_v)\phi(T_v,F_v,T_r,F_r)  \;\;\;\; \mbox{where } r \in \eta_v
\end{equation}
where $Z$ is a normalization term, $\phi$ are clique potential functions and $\eta_v$ is the set of voxels
surrounding $v$.

Segmenting a brain requires that we find the labeling $T$ with highest probability $P(T|F)$. This leads to an  optimization problem of the form $T=\arg\max_T \prod_v \phi(F_v,T_v)\phi(T_v,T_r)$ or, equivalently,
\begin{equation}
 T = \arg\min_{T \in {\cal T}} \sum_v \left ( V(F_v,T_v) + \sum_{r\in \eta_v} I(T_v,F_v,T_r,F_r) \right ). \label{eq:min}
\end{equation}
where we set the equivalence $V(F_v,T_v) = -\log \phi(F_v,T_v)$ and $I(T_v,F_v,T_r,F_r) = -\log \phi(T_v,F_v,T_r,F_r)$. 

In our case, we model the unary terms $V(F_v,T_v)$ by taking the negative log of the posterior distribution
\begin{equation}
V(F_v,T_v)=-log(P(T_v|F_v)
\end{equation}
specified in Eq.(\ref{eq:postKNN}), (\ref{eq:postSVM}) or (\ref{eq:postDT}). As for the pairwise term, we set it to be
\begin{equation}
     I(T_v,F_v,T_r,F_r) = \lambda (1-\delta(T_v,T_r))\exp\left(\frac{-\| F_v - F_r \|}{\sigma^2}\right) .
\label{eq:crf_priorterm}
\end{equation}
The choice of these unary and pairwise terms allows us to perform the optimization of Equation~\ref{eq:min} using the graphcut algorithm.

We refer to the segmentation methods using this label dependency model as {\bf kNN-CRF}, {\bf SVM-CRF}, and {\bf DT-CRF}, depending on the unary term used.

\section{Experiments}

\subsection{Experimental Setup}

All our experiments were conducted on real patient data obtained from the brain tumor segmentation challenge dataset (\citet{BRATSURL}) as part of the MICCAI conference. This dataset contains 30 patient subjects (20 high grade and 10 low grade tumors) for training and 10 (all high grade tumors) for testing. For each subject there exist 4 modalities which are co-aligned together, namely: T1, T1C, T2 and Flair . In our experiments, we used T1C, T2 and Flair only. We do not use T1 as it is not very  descriptive and using it did not improve the overall performance of the model. 
For each brain, the user is asked to manually label voxels in only two 2D slices for each class. The choice of slices depend on the size and spread of the tumor. Considering the fact that the user can choose slices from any view (i.e. axial, sagittal and coronal), the tumor coverage is sufficient and the results are not very sensitive to the slices chosen for labeling. On average, only $0.4\%$ of the voxels containing pathology and $0.03\%$ of the voxels corresponding to healthy tissue were manually selected, thus providing minimal labeled data to the algorithm. To make operations faster, we disregard all the voxels outside of the skull and consider them as healthy.

The quantitative results for each method was obtained from the BRATS online evaluation system, which provides Dice, Specificity and Sensitivity as measures of performance. These measures are defined as follows:
\begin{eqnarray}
Dice(P,T) &=& \frac{|P_1 \wedge T_1|}{(|P_1|+|T_1|)/2}, \nonumber \\
Sensitivity(P,T) &=& \frac{|P_1 \wedge T_1|}{|T_1|}, \nonumber \\
Specificity(P,T) &=& \frac{|P_0 \wedge T_0|}{|T_0|}, \nonumber
\end{eqnarray}

where $P$ represents the model predictions and $T$ represents the ground truth labels. We also note as $T_1$ and $T_0$ the subset of voxels predicted as positives and negatives for the tumor region in question. Similarly for $P_1$ and $P_0$~\citep{Menze2014}.

We report these measures for the test subjects over the three categories considered by the BRATS evaluation (i.e.\ complete, core, enhanced).  The {\em complete} category is the union of classes containing un-healthy tissue. i.e.~$\{l|l \in [ \mbox{necrosis, edema, enhancing}]\}$), the {\em core} category are classes containing tumor core {\em i.e.}~$\{l|l\in [ \mbox{necrosis, enhancing}]\}$ and  the {\em enhancing} category is the enhancing tumor class. i.e.~$\{l|l\in [ \mbox{enhancing}] \}$.
The online evaluation system also provides a ranking for every method submitted for evaluation. This includes methods from the 2013 BRATS challenge published in \citep{Menze2014} as well as anonymized unpublished methods for which no reference is available. 
The methods in each table presented in this section are ordered according to the ranking provided by the online evaluation system. 

Please note that we could not use the BRATS 2014 dataset due problems with both the system performing the evaluation and the quality of the labeled data. For these reasons the old BRATS 2014 dataset has been removed from the official website and, at the time of submitting this manuscript, the BRATS website still showed: ``Final data for BRATS 2014 to be released soon'' For these reasons, we decided to focus on the BRATS 2013 data. Also, this article does not contain any studies with human participants performed by any of the authors.


\subsection{Results and Discussion}


In this section, we report experimental results obtained with the machine learning methods presented in Section~\ref{sec:classifiers}. This includes linear SVM (LSVM), kernel SVM with rbf kernel (KSVM), our proposed product kernel SVM (PKSVM), kNN, decision trees trained with Ada-Boost (ADT), and random forests (RDT). All these methods have been explored with and without the CRF. The CRF parameters $\alpha$ and $\beta$ were set for each method, by cross-validation on 6 brains on the training set.  We also investigate the extent to which adding spatial features $\langle i,j,k\rangle$  helps improving the performance. This is noted by adding a ``$*$" next to the method's name.


\subsubsection{KNN}
The results for the kNN related experiments are presented in Table \ref{tab:knn}. We first made an experiment without including the $\langle i,j,k\rangle$ position features in the feature vector as presented by \cite{vaidyanathan1995}. Since his method uses neither the spatial coordinate features nor the CRF regularization, it performs significantly worse than other kNN related experiments. While adding the spatial coordinates to this method improves the result by a significant margin, the best performance is achieved when we  use both spatial coordinates and a CRF regularization.

\begin{table*}[tp]
\caption{Dice, Specificity and Sensitivity measures for kNN methods on BRATS-2013 test set.
}
\begin{center}
\resizebox{\textwidth}{!}{%
\begin{tabular}{*{10}{c}}
\hline
Method \multirow{2}*{ }&\multicolumn{3}{c}{Dice}&\multicolumn{3}{c}{Specificity }&\multicolumn{3}{c}{Sensitivity}\\
\cline{2-10}
  &Complete &Core &Enhancing &Complete &Core &Enhancing &Complete &Core &Enhancing\\ \hline
kNN-CRF*  & 0.85 & 0.75 & 0.60 & 0.91 & 0.85 & 0.77 & 0.78 & 0.69 & 0.56\\
kNN*      & 0.81 & 0.68 & 0.65 & 0.76 & 0.62 & 0.62 & 0.90 & 0.84 & 0.73\\
kNN-CRF   & 0.80 & 0.69 & 0.55 & 0.92 & 0.83 & 0.75 & 0.74 & 0.63 & 0.48\\
kNN       & 0.65 & 0.52 & 0.53 & 0.59 & 0.49 & 0.50 & 0.77 & 0.68 & 0.65\\
\hline
\end{tabular}
}
\end{center}
\label{tab:knn}
\end{table*}

\subsubsection{SVM}

The results for the SVM-related experiments are presented in Table~\ref{tab:svm_results}. Results confirm that using spatial coordinate features (shown with "*") and using the CRF model (shown with "-CRF") improve the performance of both a linear SVM (LSVM) and an RBF kernel SVM (KSVM). 
It is also quite clear from this experiment that the non-linearity of the kernel SVM is crucial, as it significantly outperforms the linear SVM (LSVM). 

As for the PKSVM method which stands for the RBF product kernel SVM presented in Section~\ref{sec:distance} (c.f. Eq.(\ref{eq:kernel-non-lin})) it clearly improved the Kernel-SVM and Kernel-SVM+CRF results.  This underlines the relative importance of the spatial coordinate features $\langle i,j,k\rangle$ versus the input T1, T2 and Flair modalities. 

\begin{table*}[tp]
\caption{Dice, Specificity and Sensitivity measures for various SVM methods on the BRATS-2013 test set.
}
\begin{center}
\resizebox{\textwidth}{!}{%
\begin{tabular}{*{10}{c}}
\hline
Method \multirow{2}*{ }&\multicolumn{3}{c}{Dice}&\multicolumn{3}{c}{Specificity }&\multicolumn{3}{c}{Sensitivity}\\
\cline{2-10}
  &Complete &Core &Enhancing &Complete &Core &Enhancing &Complete &Core &Enhancing\\ \hline
PKSVM-CRF* & 0.86 & 0.77 & 0.73 & 0.88 & 0.85 & 0.76 & 0.78 & 0.68 & 0.58\\
KSVM-CRF*  & 0.84 & 0.75 & 0.70 & 0.87 & 0.77 & 0.72 & 0.82 & 0.79 & 0.71\\
PKSVM*     & 0.82 & 0.71 & 0.69 & 0.84 & 0.73 & 0.71 & 0.80 & 0.76 & 0.71\\
KSVM*      & 0.81 & 0.68 & 0.65 & 0.76 & 0.62 & 0.62 & 0.90 & 0.84 & 0.73\\
KSVM-CRF   & 0.74 & 0.67 & 0.53 & 0.82 & 0.82 & 0.79 & 0.73 & 0.61 & 0.45\\
LSVM-CRF*  & 0.79 & 0.64 & 0.51 & 0.86 & 0.74 & 0.70 & 0.74 & 0.62 & 0.45\\
LSVM*      & 0.69 & 0.59 & 0.62 & 0.65 & 0.54 & 0.47 & 0.84 & 0.76 & 0.59\\
LSVM-CRF   & 0.72 & 0.60 & 0.46 & 0.77 & 0.66 & 0.59 & 0.72 & 0.61 & 0.44\\
KSVM       & 0.65 & 0.50 & 0.50 & 0.61 & 0.49 & 0.49 & 0.75 & 0.63 & 0.58\\
LSVM       & 0.51 & 0.35 & 0.45 & 0.48 & 0.35 & 0.43 & 0.73 & 0.59 & 0.59\\
\hline
\end{tabular}
}
\end{center}
\label{tab:svm_results}
\end{table*}

\subsubsection{Decision trees}

For these experiments, we fixed the number of decision trees for AdaBoost (ADT) and random forests (RDT) to 100 and the leaf size to $1$. For AdaBoost, decision stumps were used. The quantitative results are shown in Table~\ref{tab:tree_result}. While adding spatial features are beneficial for both random forests and AdaBoost, using the CRF model is mostly beneficial except for random forest without spatial coordinates. However, the segmentation systems relying on decision trees tend to be worse than using kNN or SVM methods.

\begin{table*}[tp]
\caption{Dice, Specificity and Sensitivity measures for ensemble of decision trees with AdaBoost (ADT) and random forests (RDT) on BRATS-2013 test dataset.
}
\begin{center}
\resizebox{\textwidth}{!}{%
\begin{tabular}{*{10}{c}}
\hline
Method \multirow{2}*{ }&\multicolumn{3}{c}{Dice}&\multicolumn{3}{c}{Specificity }&\multicolumn{3}{c}{Sensitivity}\\
\cline{2-10}
  &Complete &Core &Enhancing &Complete &Core &Enhancing &Complete &Core &Enhancing\\ \hline
RDT*     & 0.81 & 0.69 & 0.64 & 0.83 & 0.71 & 0.64 & 0.79 & 0.75 & 0.70\\
RDT-CRF* & 0.82 & 0.69 & 0.51 & 0.92 & 0.83 & 0.79 & 0.73 & 0.61 & 0.50\\
RDT-CRF  & 0.80 & 0.66 & 0.49 & 0.92 & 0.83 & 0.78 & 0.71 & 0.60 & 0.40\\
ADT-CRF* & 0.79 & 0.64 & 0.51 & 0.88 & 0.75 & 0.71 & 0.72 & 0.61 & 0.45\\
ADT-CRF  & 0.78 & 0.63 & 0.50 & 0.87 & 0.73 & 0.67 & 0.72 & 0.61 & 0.45\\
ADT*     & 0.73 & 0.57 & 0.58 & 0.73 & 0.60 & 0.59 & 0.75 & 0.64 & 0.66\\
RDT      & 0.67 & 0.55 & 0.55 & 0.66 & 0.55 & 0.53 & 0.72 & 0.65 & 0.65\\
ADT  	 & 0.65 & 0.48 & 0.54 & 0.66 & 0.55 & 0.53 & 0.69 & 0.52 & 0.62\\

\hline	

\end{tabular}
}
\end{center}
\label{tab:tree_result}
\end{table*}



\subsubsection{Robustness of hyper-parameter selection}
\label{sec:hyper-parameter_selection}


In our method when using the SVM as the classifier, the hyper-parameters (regularization constant $C$ and kernel hyper-parameters $\gamma$, $\gamma_1$ and $\gamma_2$) were always cross-validated for each brain individually, using an automated grid search. On the other hand, for automatic methods, a fixed set of hyper-parameters is used for generalization. Given the variation of the MRI data and tumor types, we hypothesize that using a fixed set of hyper-parameters will degrade the performance quite significantly. 

To evaluate the importance of performing per-brain model selection, we conducted an experiment  where we used a fixed configuration of hyper-parameters for all subjects. For this experiment, we considered our top two segmentation methods, PKSVM-CRF* and KSVM-CRF*. The values of the hyper-parameters were chosen by taking the hyper-parameter value most frequently selected by these methods, across all the brains. The idea was to pick values that are most likely to work well in general. 
For the KSVM-CRF*, $C$ was set to $1$ and $\gamma$ to $5$ and for the PKSVM-CRF*, $C$ was set to $1$, $\gamma_1$ to 100 and $\gamma_2$ to 10.    

The results (Table~\ref{tab:shared_svm}) show a decrease in performance if fixed hyper-parameters are used for all brains. We also performed this experiment on the BRATS training data (not shown here) and the performance decreased even more. This was not unexpected, since the training data is more varied and actually consists of both high grade tumors and low grade tumors, while the test data only contains high grade tumors. 

\begin{figure}[ht] 
\centering
\includegraphics[scale=0.37]{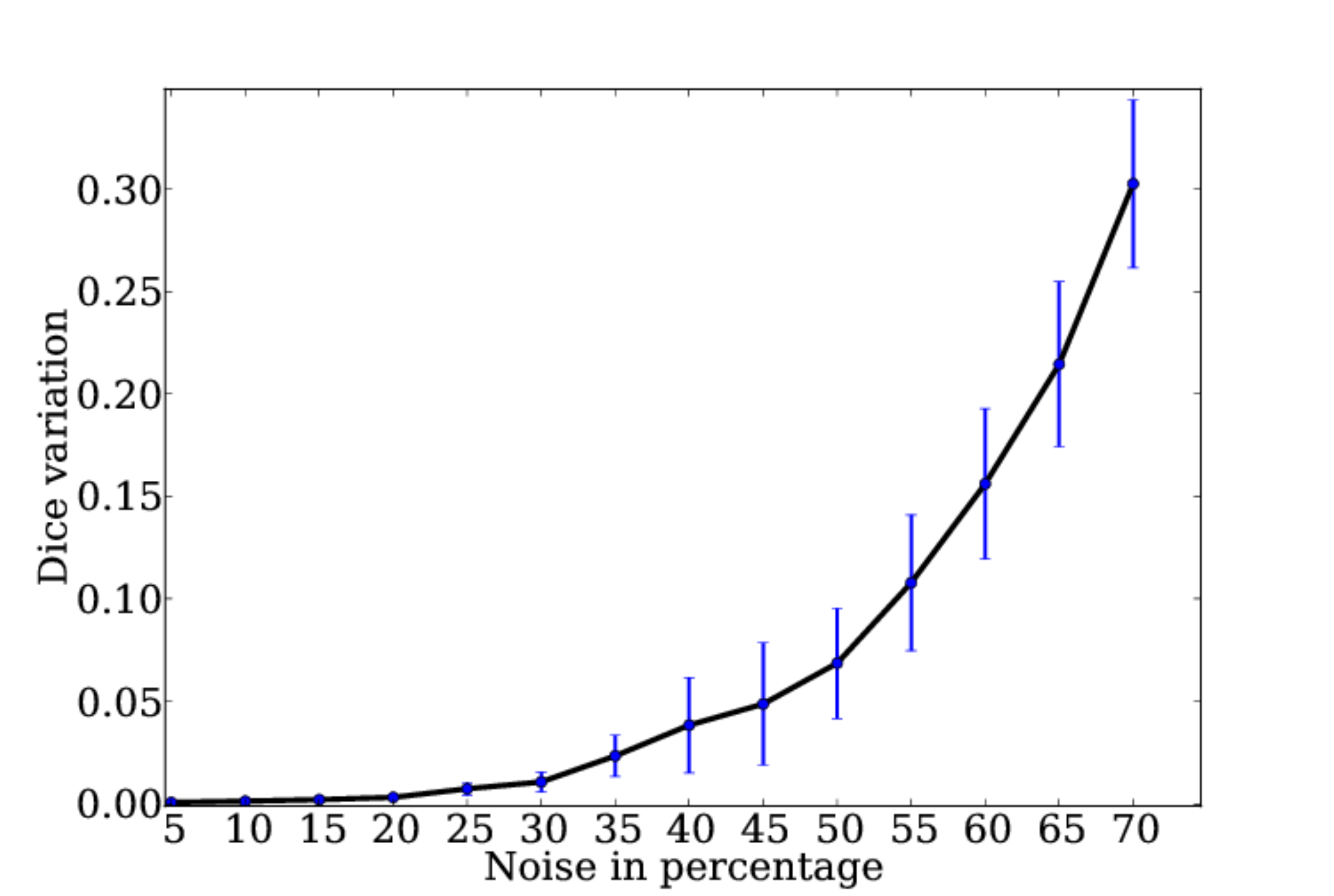}

\caption{Sensitivity of the model with respect to the gamma hyper parameter. }
\label{fig:traf}
\end{figure}

While it appears the tuning of the SVM's hyper-parameter to each brain is beneficial, we tested the extent to which small changes to the optimal hyper-parameters would affect the performance. This is meant to simulate the fact that cross-validation might not always find the same hyper-parameters between variations on the manually labeled voxels.  In order to measure how resilient our method is to slight hyper-parametric shifts, we ran another experiment to measure the sensitivity of our model.  We did so by randomly selecting 20 brains from the BRATS training data, trained an SVM whose hyper-parameters have been obtained from cross validation.  We then added noise to the hyper-parameters and measured the effect on the resulting segmentation. The noise corresponded to Gaussian noise, whose standard deviation was set to a certain percentage of the hyper-parameters' values. Figure~\ref{fig:traf} shows the resulting Dice measure for different noise level. As one can see, even with a noise level corresponding to a corruption of $25\%$ of the hyper-parameter values, the end result is still close to the one obtained without any noise. 

Finally, the importance of optimizing the hyper-parameters was found to be less crucial for the other methods. For kNN, we evaluated the effect of using different values of $k$, with $k = 3$ consistently producing higher performance.  The same type of experiment was performed to measure the effect of using different number of trees and leaf size in ADT and RDT.  For these methods, setting the number of decision trees to $100$ and leaf size to $1$ always worked well. 

\begin{table*}[tp!]
\caption{The effect of having a fixed selection of hyper-parameters for kernel SVM and product kernel SVM.}
\begin{center}
\resizebox{\textwidth}{!}{%
\begin{tabular}{*{10}{c}}
\hline
Method \multirow{2}*{ }&\multicolumn{3}{c}{Dice}&\multicolumn{3}{c}{Specificity }&\multicolumn{3}{c}{Sensitivity}\\
\cline{2-10}
  &Complete &Core &Enhancing &Complete &Core &Enhancing &Complete &Core &Enhancing\\ \hline
PKSVM-CRF*     & 0.86 & 0.77 & 0.73 & 0.88 & 0.85 & 0.76 & 0.78 & 0.68 & 0.58\\
KSVM-CRF*      & 0.84 & 0.75 & 0.70 & 0.87 & 0.77 & 0.72 & 0.82 & 0.79 & 0.71\\
FixedKSVM-CRF* & 0.82 & 0.69 & 0.56 & 0.93 & 0.82 & 0.78 & 0.75 & 0.64 & 0.49\\
FixedPSVM-CRF* & 0.72 & 0.56 & 0.55 & 0.71 & 0.62 & 0.58 & 0.73 & 0.65 & 0.65\\
\hline
\end{tabular}
}
\end{center}
\label{tab:shared_svm}
\end{table*}

\subsubsection{Speed-up procedure}

\begin{figure}[htp] 
\centering
\subfloat[]{\label{fig:traf2-a}\includegraphics[height = 6 cm]{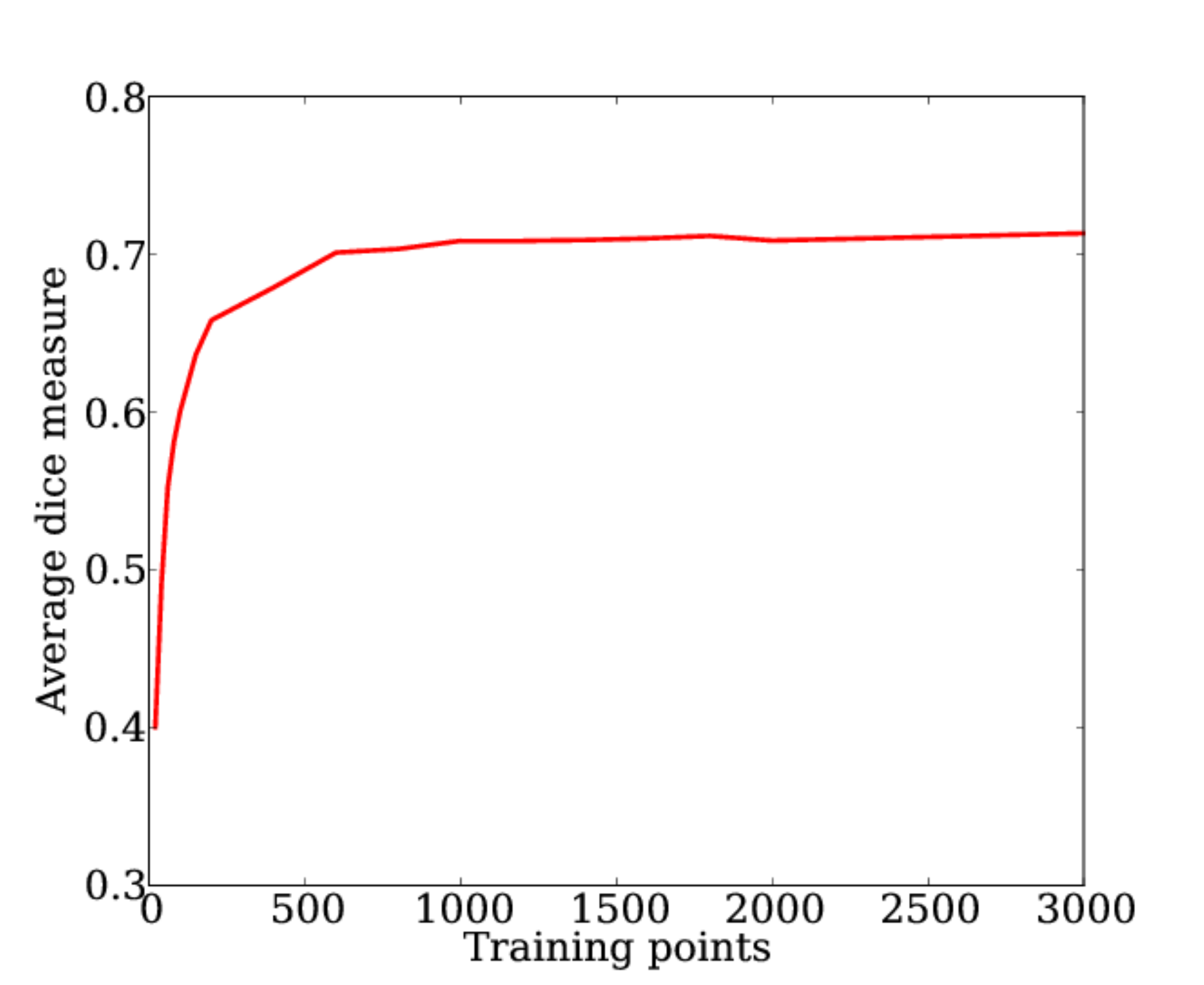}}
\subfloat[]{\label{fig:traf2-b}\includegraphics[height = 6 cm]{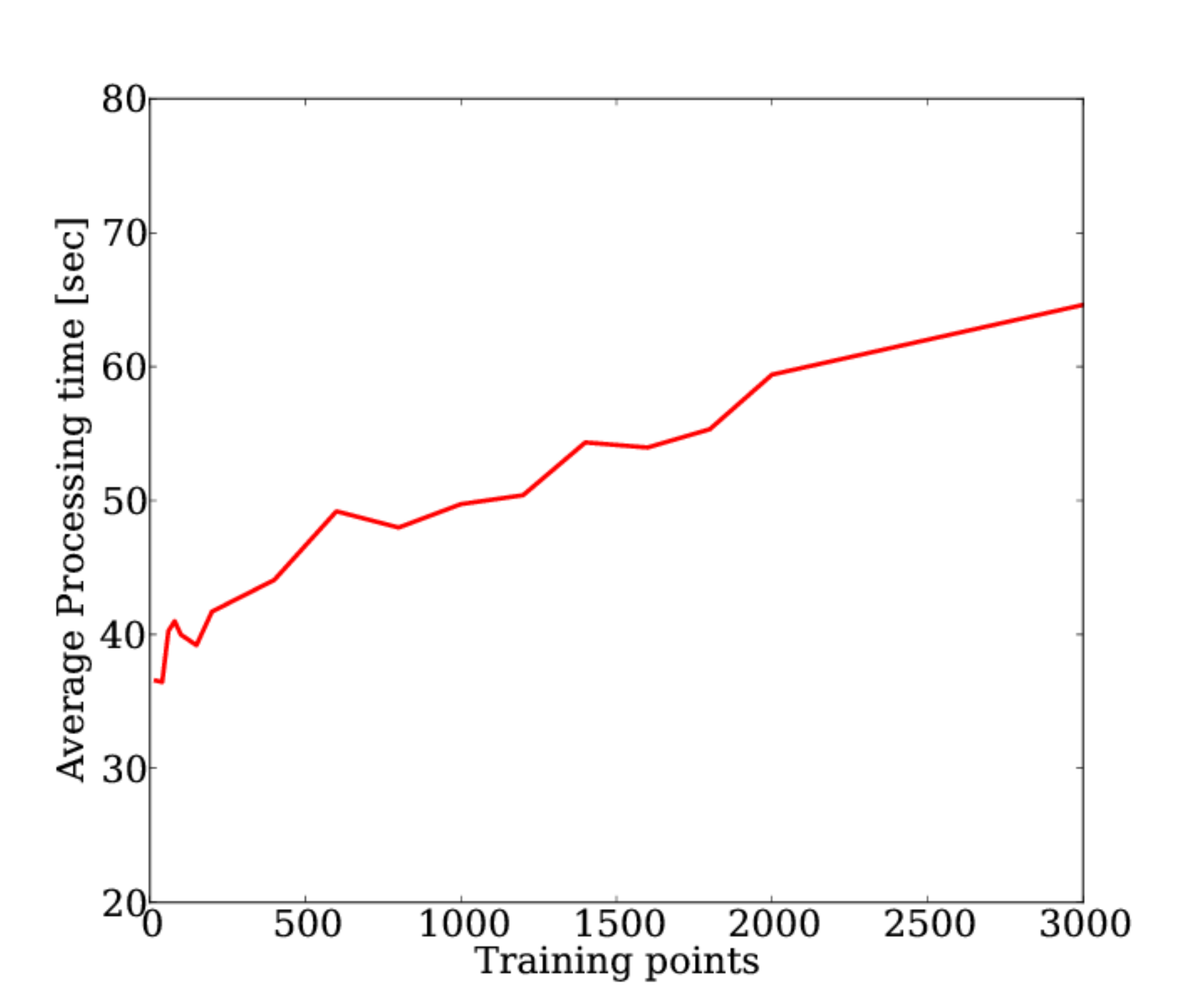}}
\caption{Sensitivity of the model with respect to the number of training points. (a) shows variation in average Dice measure while (b) shows variation in the average processing time and memory usage. }
\label{fig:traf2}
\end{figure}

Every segmentation method presented in this paper uses manually-selected voxels as their input.  However, these selected voxels often carry out similar information.  That is especially true for neighboring voxels whose  $\langle i,j,k  \rangle$ position is almost the same, and whose T1,T2, Flair values are likely to be identical.  Thus, in order to speed-up the segmentation procedure, one can randomly down-sample the training data.  To have an overall idea to what extent we can down-sample the data without hurting too much the overall precision, we conducted an experiment where we divide the training points into healthy and non-healthy subsets and subsample them separately.  This is done as to maintain a balance between the size of the healthy class with respect to other classes. The outcome of this process is a smaller training set but with roughly the same proportion of healthy points and non-healthy points. 
Figure~\ref{fig:traf2} shows the result of this experiment. The curves were obtained by averaging the results of 20 randomly selected brains from BRATS training data. The horizontal axes in Figure~\ref{fig:traf2} shows the number of training points in the subsampled training set.
As shown in Figure \ref{fig:traf2}(a), 
with maximum number of training points (i.e 3000) we get an average Dice measure of $0.72$ and by considering $1000$ training points the average Dice measure barely drops to $0.71$, while the processing time decreases by 60\%.  
Thus, all experiments submitted to the BRATS website were done with this subsampling measure. 

\begin{figure*}[ht!]
    \begin{center}
        \includegraphics[width=0.85\linewidth]{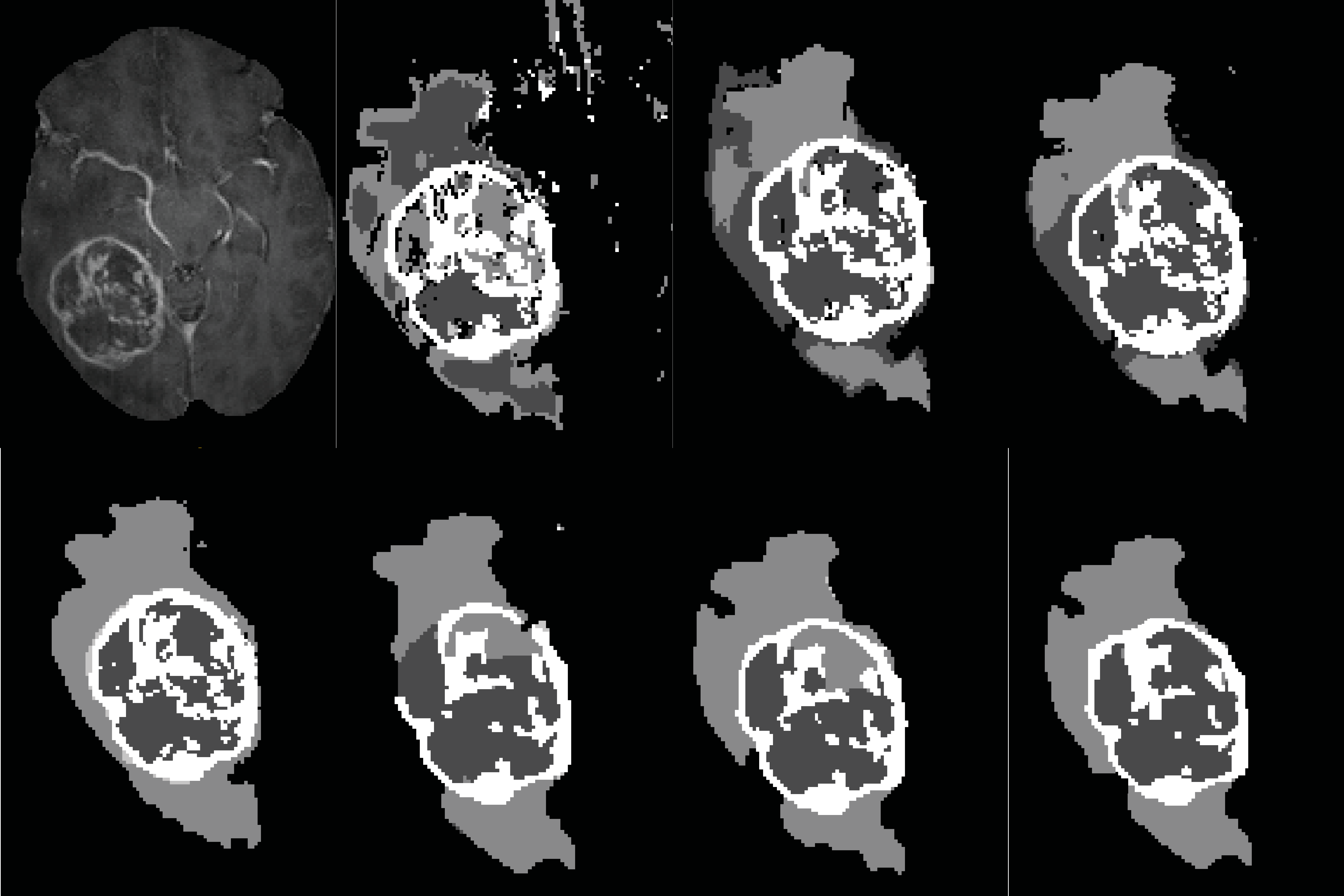}
	    \caption{Illustration of brain tumor segmentation maps predicted by different variations of SVM. Top row from left to right : T1C modality, KSVM, KSVM*, PKSVM*. Bottom row from left to right: ground truth, KSVM-CRF, KSVM*-CRF, PKSVM*-CRF.\vspace{-0.4cm}
	        }
	    \label{fig:segmentations} 
	\end{center}
\end{figure*}

\vspace{-0.2cm}
\section{Conclusion}
\vspace{-0.2cm}

\subsection{Putting it all together}


We finally present how our top performing methods compare with other state-of-the-art methods. The BRATS official website provides a ranking system for this purpose. However, because the BRATS organizers have recently made all methods anonymous, a complete comparison is not possible. For that reason, we rank our method based on the MICCAI-BRATS 2013 challenge results for which references to the methods were available. This is shown in table~\ref{tab:bestmethod} \footnote{Please note that the results mentioned in Table~\ref{tab:bestmethod} are from methods competing in the BRATS 2013 challenge for which a static table is provided [\text{https://www.virtualskeleton.ch/BRATS/StaticResults2013}]. Since then, other methods have been added to the score board but for which no reference is available.}.  As one can see, PKSVM-CRF* and KSVM-CRF* are ranked second and third respectively, closely behind Tustison et al.\ and kNN-CRF* is ranked $6^{th}$ in this table. 
Using the spatial features $\langle i,j,k\rangle$, and CRF post-processing is vital to produce highly accurate results. 
Many methods in this table (like that of Tustison et al.\, Reza et al.\ and Festa et al.\ ) use random forests with a large number of features. In our case, random forests did not perform as well as the SVM or kNN methods. This might be due to the low dimensionality of our feature space. Recently \citet{Subbanna2014} published competitive results on the BRATS 2013 dataset, reporting Dice measures of $0.86, 0.86, 0.77$ for Complete, Core and Enhancing tumor regions. Since they do not report Specificity and Sensitivity measures, a completely fair comparison with that method is not possible. However, as mentioned in~\citep{Subbanna2014}, their method takes 70 minutes to process a subject, which is significantly slower than our method. 

Figure~\ref{fig:segmentations} shows 
a visualisation of segmentation results, for different variations of our SVM method. 
This illustrates the contribution of adding spatial features, using a CRF and using our improved kernel function, in improving the general performance of the SVM approach.  

 
 \begin{table*}[tp]
\caption{Comparison of our top implemented architectures with the state-of-the-art methods on the BRATS-2013 test set.
}
\begin{center}
\resizebox{\textwidth}{!}{%
\begin{tabular}{*{10}{c}}
\hline
Method \multirow{2}*{ }&\multicolumn{3}{c}{Dice}&\multicolumn{3}{c}{Specificity }&\multicolumn{3}{c}{Sensitivity}\\
\cline{2-10}
  &Complete &Core &Enhancing &Complete &Core &Enhancing &Complete &Core &Enhancing\\ \hline
Tustison	&$0.87$	&$0.78$	&$0.74$	&$0.85$	&$0.74$	&$0.69$	&$0.89$	&$0.88$	&$0.83$	\\ 
 \rowcolor[gray]{0.9}PKSVM-CRF* &  0.86  &  0.77  &  0.73   &0.88       &0.85 &	0.76&	0.78 &	0.68 &	0.58  \\ 
 \rowcolor[gray]{0.9}KSVM-CRF* &0.84  &	0.75  &	0.70  &	0.87  &	0.77  &	0.72  &	0.82  &	0.79  &	0.71  \\
 \rowcolor[gray]{0.9}kNN-CRF* &	0.85 &	0.75&	0.60 &	0.91&	0.85&	0.77&	0.78 &	0.69 &	0.56\\
Meier    	&$0.82$	&$0.73$	&$0.69$	&$0.76$	&$0.78$	&$0.71$	&$0.92$	&$0.72$	&$0.73$\\ 
Reza  	&$0.83$	&$0.72$	&$0.72$	&$0.82$	&$0.81$	&$0.70$	&$0.86$	&$0.69$	&$0.76$	\\ 
Zhao    	&$0.84$	&$0.70$	&$0.65$	&$0.80$	&$0.67$	&$0.65$	&$0.89$	&$0.79$	&$0.70$\\ 
Cordier 	&$0.84$	&$0.68$	&$0.65$	&$0.88$	&$0.63$	&$0.68$	&$0.81$	&$0.82$	&$0.66$\\ 
Festa 	&$0.72$	&$0.66$	&$0.67$	&$0.77$	&$0.77$	&$0.70$	&$0.72$	&$0.60$	&$0.70$	\\ 
Doyle 	&$0.71$	&$0.46$	&$0.52$	&$0.66$	&$0.38$	&$0.58$	&$0.87$	&$0.70$	&$0.55$	\\ \hline
\end{tabular}
}
\end{center}
\label{tab:bestmethod}
\end{table*}
\subsection{Processing time and memory usage}
A key advantage of our proposed method is in having a very small processing time and memory usage, while maintaining high accuracy. Due to the low dimensionality of our feature space, it only takes up, on average, 50 $\mathrm{MB}$ of RAM to store the feature space of a brain. This is very small compared to state-of-the-art methods, whose memory footprint of the feature space is on the order of GB's. For example, Festa et al.\ use a feature space of 300 dimensions for their random forest approach which would take up to 2.7GB's. Tustison et al.\, Reza et al.\ and Meier et al.\ also take a similar approach using random forests~\citep{Menze2014}. These methods rely on a high number of texture features which are computationally time consuming and memory wise expensive. 

Apart from the feature space, our proposed methods have different speed and memory footprint. We can make a comparison in accuracy, speed and memory usage as presented in Table~\ref{tab:bestofOurMethods}. The processing time was measured on an 8-core processor and includes both training and testing. The time required by graphcut inference is the same for all methods and involves only an additional 8 seconds. As shown in Table~\ref{tab:bestofOurMethods}, PKSVM-CRF* has the highest accuracy but requires a higher processing time (35 seconds) and memory usage (7.7 MB), on top of the 50 MB required to store the feature space. On the other hand, KSVM-CRF* and kNN-CRF* are closer to real time implementations with negligeable memory consumption. This allows the expert to interact in real-time with the software. That being said, all methods presented in Table~\ref{tab:bestofOurMethods} are significantly faster than state-of-the-art methods. For example, Tustison's method takes around 30 minutes to process a brain as mentioned in \citet{Menze2014}.

\begin{table*}[tp]
\caption{Best performing methods for each machine learning category with average processing time and memory usage.}
\begin{center}
\resizebox{\textwidth}{!}{%
\begin{tabular}{*{12}{c}}
\hline
Method \multirow{2}*{ }&\multicolumn{3}{c}{Dice}&\multicolumn{3}{c}{Specificity }&\multicolumn{3}{c}{Sensitivity} &Time &Memory\\
\cline{2-10}
  &Complete &Core &Enhancing &Complete &Core &Enhancing &Complete &Core &Enhancing\\ \hline
PKSVM-CRF* & 0.82 & 0.71 & 0.69 & 0.84 & 0.73 & 0.71 & 0.80 & 0.76 & 0.71 & 35sec & 7.7$\mathrm{MB}$\\
KSVM-CRF*  & 0.81 & 0.68 & 0.65 & 0.76 & 0.62 & 0.62 & 0.90 & 0.84 & 0.73 & 10sec & 75$\mathrm{KB}$\\
kNN-CRF*   & 0.81 & 0.68 & 0.65 & 0.76 & 0.62 & 0.62 & 0.90 & 0.84 & 0.73 & 3sec. & 40$\mathrm{KB}$\\
RDT*   & 0.81 & 0.69 & 0.64 & 0.83 & 0.71 & 0.64 & 0.79 & 0.75 & 0.70 & 10sec & 120$\mathrm{KB}$\\
\hline

\end{tabular}
}
\end{center}
\label{tab:bestofOurMethods}
\end{table*}

In this paper we evaluated the capability of {\it within brain generalization} using a variety of classifiers.
We showed that the SVM reached the best performances, thanks in part to a kernel function specifically adapted to our feature space. 
Most interestingly, we also showed that adopting a fixed hyper-parameter configuration for all brains actually decreases the performance of the SVM. A better strategy was to also perform hyper-parameter selection for each brain individually, in order to adapt to the specificities of each brain, further motivating our {\it within brain generalization} framework. 

\section{Conflict of Interest}
The authors declare that they have no conflict of interest.
\section{Ethical approval}
All procedures performed in studies involving human participants were in accordance with the ethical standards of the institutional and/or national research committee and with the 1964 Helsinki declaration and its later amendments or comparable ethical standards.

This article does not contain any studies with human participants performed by any of the authors.





 \bibliographystyle{spbasic} 

\bibliography{strings,reference}

\end{document}